\documentclass[letterpaper]{article} % DO NOT CHANGE THIS

\usepackage{aaai24}

\usepackage{times}  % DO NOT CHANGE THIS
\usepackage{helvet}  % DO NOT CHANGE THIS
\usepackage{courier}  % DO NOT CHANGE THIS
\usepackage[hyphens]{url}  % DO NOT CHANGE THIS
\usepackage{graphicx} % DO NOT CHANGE THIS
\usepackage{natbib}  % DO NOT CHANGE THIS AND DO NOT ADD ANY OPTIONS TO IT
\usepackage{caption} % DO NOT CHANGE THIS AND DO NOT ADD ANY OPTIONS TO IT
\usepackage{subcaption}
\usepackage{graphicx}
\usepackage{booktabs}
\usepackage{amssymb}
\usepackage{amsmath}
\usepackage{gensymb}
\usepackage{algorithm}
\usepackage{algpseudocode}
\usepackage{accents} 
\usepackage{mdframed}
\usepackage{multirow}
\usepackage{adjustbox}

\newcommand*{\vv}[1]{\vec{\mkern0mu#1}}
\newcommand{\act}[1]{{\tt\MakeUppercase{#1}}}
\DeclareMathOperator*{\argminA}{arg\,max}

\usepackage{newfloat}
\usepackage{listings}
\DeclareCaptionStyle{ruled}{labelfont=normalfont,labelsep=colon,strut=off} % DO NOT CHANGE THIS
\floatstyle{ruled}
\newfloat{listing}{tb}{lst}{}
\floatname{listing}{Listing}
\title{VELMA: Verbalization Embodiment of LLM Agents for Vision and Language Navigation in Street View}
\author {
    Raphael Schumann\textsuperscript{\rm 1},
    Wanrong Zhu\textsuperscript{\rm 2},
    Weixi Feng\textsuperscript{\rm 2},
    Tsu-Jui Fu\textsuperscript{\rm 2},\\
    Stefan Riezler\textsuperscript{\rm 1,3},
    William Yang Wang\textsuperscript{\rm 2}
}
\affiliations {
    \{\textsuperscript{\rm 1}Computational Linguistics, \textsuperscript{\rm 3}IWR\}, Heidelberg University, Germany\\
    \texttt{\normalsize \{rschuman, riezler\}@cl.uni-heidelberg.de}\\
    \textsuperscript{\rm 2}University of California, Santa Barbara\\
    \texttt{\normalsize \{wanrongzhu, weixifeng, tsu-juifu, william\}@cs.ucsb.edu}
}

\begin{document}

\maketitle

\begin{abstract}
Incremental decision making in real-world environments is one of the most challenging tasks in embodied artificial intelligence. One particularly demanding scenario is Vision and Language Navigation~(VLN) which requires visual and natural language understanding as well as spatial and temporal reasoning capabilities. The embodied agent needs to ground its understanding of navigation instructions in observations of a real-world environment like Street View. Despite the impressive results of LLMs in other research areas, it is an ongoing problem of how to best connect them with an interactive visual environment. In this work, we propose VELMA, an embodied LLM agent that uses a verbalization of the trajectory and of visual environment observations as contextual prompt for the next action. Visual information is verbalized by a pipeline that extracts landmarks from the human written navigation instructions and uses CLIP to determine their visibility in the current panorama view. We show that VELMA is able to successfully follow navigation instructions in Street View with only two in-context examples. We further finetune the LLM agent on a few thousand examples and achieve around 25\% relative improvement in task completion over the previous state-of-the-art for two datasets.
\end{abstract}

\section{Introduction}
\begin{figure}[!t]
    \centering
    \includegraphics[width=0.49\textwidth]{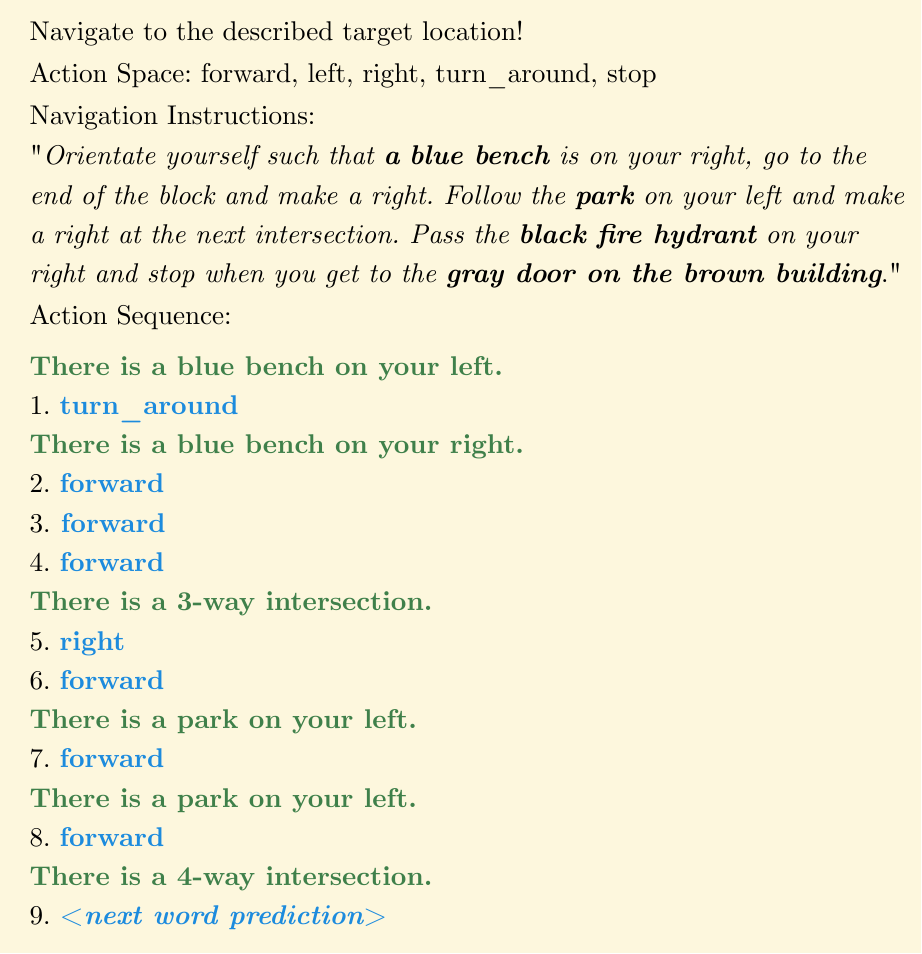}
    \caption{Prompt sequence used to utilize LLMs for VLN in Street View. Verbalized observations of the visual environment are in green and appended to the prompt at each step. Agent actions~(blue) are acquired by LLM next word prediction. Highlighting of text for visual presentation only. Full navigation trajectories are, on average, 40 steps long.}
    \label{fig:figure1}
\end{figure}

Large language models (LLMs), which have shown impressive reasoning capabilities in traditional natural language processing tasks, are increasingly used as the reasoning engine of embodied agents for, e.g., household robots~\cite{ALFRED20}, video games~\cite{wang2023voyager} and indoor navigation~\cite{zhou2023navgpt}. These tasks are mostly based on simulations that either feature computer-generated images with a fixed set of displayable objects and textures, or are limited in scale and trajectory length. In this paper, we present a verbalization embodiment of an LLM agent~(VELMA) for urban vision and language navigation in Street View. The unique challenge of this task is the combination of a large-scale environment derived from an actual road network, real-world panorama images with dense street scenes, and long navigation trajectories. The agent needs to ground its understanding of the navigation instructions in the observable environment and reason about the next action to reach the target location. The navigation instructions are written by humans and include open-ended landmark references and directional indications intended to guide the agent along the desired path. In order to leverage the reasoning capabilities of LLMs, we use embodiment by verbalization, a workflow where the task, including the agent's trajectory and visual observations of the environment, is verbalized, thus embodying the LLM via natural language. Figure~\ref{fig:figure1} shows the verbalization at the ninth step of the current trajectory for a given navigation instance. At each step, the LLM is prompted with the current text sequence in order to predict the next action. Then the predicted action is executed in the environment, and the new observations are verbalized and appended to the prompt. This is repeated until the agent eventually predicts to stop. 

The main contributions of our work are as follows: (i)~We introduce VELMA, to our knowledge, the first LLM-based agent for urban VLN. (ii)~We report few-shot results for the urban VLN task and achieve new state-of-the-art performance by finetuning our agent on the training set. (iii)~We address and resolve limitations of the commonly used Touchdown environment~\cite{Chen2018Touchdown}, making it amenable for few-shot agents.

\section{Related Work}
\paragraph{Outdoor VLN}
Agent models for the outdoor/urban VLN task~\cite{Chen2018Touchdown} commonly follow a sequence-to-sequence architecture where encoded text and image representations are fused for each decoder step~\cite{Xiang2020LearningNavigation, directions_in_streetview, Mehta2020Retouchdown, schumann-riezler-2022-analyzing, s23136028}. Other proposed agents employ pretrained vision and language transformers that are finetuned on task-specific data~\cite{zhu-etal-2021-multimodal, armitage2023priority}. \citet{NEURIPS2021_b3e3e393} represent the visual environment by symbols using semantic segmentation and extreme downsampling of panorama images, but their agent does not improve over previous success rates. Other work uses CLIP to score the presence of extracted landmarks at each panorama node in a graph and uses this information to plan a route for given navigation instructions~\cite{shah2022lmnav}. Their non-urban environment has a graph with 300 nodes, and the navigation path is planned a priori with full access to all panorama images and landmark scores. In contrast, our agent is embodied and has to plan ad-hoc with access to directly observed information only.
\paragraph{Indoor VLN}
Indoor agents~\cite{fried2018speaker, Wang_2019_CVPR, tan-etal-2019-learning, fu2020counterfactual, zhu2020babywalk, qi2020reverie, hong2021vln, chen2021history, li2022envedit} are used for navigation datasets like R2R~\cite{anderson2018vision} and RxR~\cite{ku2020room} or ObjectNav~\cite{ramakrishnan2021habitatmatterport, zhou2023esc}. \citet{Khandelwal_2022_CVPR} showed that using the CLIP encoder for image features improves performance for a range of vision and language tasks. Recently, \citet{zhou2023navgpt} introduced an LLM-based agent for R2R that incorporates image information by transcribing its entire content with an image-to-text model. This is feasible because the navigation trajectories are only six steps on average compared to 40 steps in the urban VLN task considered in our work. Another notable indoor VLN agent~\cite{Dorbala2022} uses CLIP to directly predict the next action by scoring the compatibility of a sub-instruction with available waypoint images.

\section{Urban VLN Environment}
\label{sec:env}
We use the Touchdown environment introduced by~\citet{Chen2018Touchdown}. The environment is based on Google's Street View and features 29,641 full-circle panorama images connected by a navigation graph. It covers the dense urban street network spanning lower Manhattan. The navigation graph is a directed graph $G=\langle V,E \rangle$ where each edge $\langle v,v' \rangle \in E$ is associated with $\alpha_{\langle v,v' \rangle}$ which is the heading direction from node $v$ to node $v'$ ranging from 0\degree~to 360\degree. The agent state $s=(v, \alpha)$ is composed of its current position $v \in V$ and its heading direction $\alpha$. The agent can move by executing an action $a \in \{\act{forward},\act{left},\act{right},\act{stop}\}$. The state transition function $s_{t+1} = \phi(a_t, s_t)$ defines the behavior of the agent executing an action. In \citet{Chen2018Touchdown}, the agent's heading $\alpha_t$ at position $v$ is restricted to align with the heading of an outgoing edge $\alpha_{\langle v,v' \rangle}$. In case of the $\act{right}$ action, the new state $s_{t+1}$ is $(v,\alpha_{\langle v,\vv{v} \rangle})$ where $\vv{v}$ is the neighboring node closest to the right of the agent's current heading. In other words, the agent is rotated in place to the right until it \textit{snaps} to the direction of an outgoing edge. Likewise, for the $\act{left}$ action. In the case of the $\act{forward}$ action, the agent moves along the edge $\langle v,v' \rangle$ according to its current heading direction $\alpha_{t} = \alpha_{\langle v,v' \rangle}$. The environment is then forced to automatically rotate the agent's heading towards an outgoing edge: $\alpha_{t+1} = \alpha_{\langle v',v^* \rangle}$ where $v^*$ is the neighbor node in the direction closest to the previous heading $\alpha_{t}$.

\subsection{Alignment Inconsistencies in Touchdown}
\label{sec:inconsistencies}
\begin{figure}[t]
    \centering
    \includegraphics[width=0.98\linewidth]{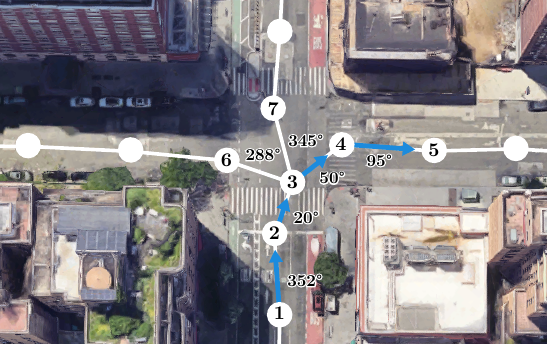}
    \caption{The Touchdown environment introduced by \citet{Chen2018Touchdown} can require action sequences that are semantically inconsistent with the correct navigation instructions. In the depicted subgraph, the action sequence to move from node~1 to node~5 is to move $\act{forward}$ four times. The semantically correct sequence of actions would include a right turn in between. We fix the problem by modifying the environment behavior and selecting the desired direction at intersections in relation to all outgoing streets.}
    \label{fig:intersection}
\end{figure}

As described in \citet{schumann-riezler-2022-analyzing}, the automatic rotation mentioned above can lead to generalization problems, e.g., when moving towards the flat side of a T-intersection. For example, if the agent is automatically rotated towards the right facing street and subsequently executes the $\act{right}$ action, it rotates towards the direction it came from instead of clearing the intersection in the intended direction. The same problem also occurs at intersections with more than three directions. Figure~\ref{fig:intersection} gives an illustrative example that shows the navigation graph at a 4-way intersection. Because the environment is derived from a real-world street layout, the nodes in the graph are not perfectly arranged as in an artificial grid world. In order to make a right turn at the intersection and to follow the route from $v^1$ to $v^5$, one expects to use the action sequence $[\act{forward}, \act{forward}, \act{right}, \act{forward}, \act{forward}]$. However, when the agent reaches $v^3$, it is automatically rotated towards the closest outgoing edge, in this case, $\langle v^3,v^4 \rangle$. This is because the rotation $20\degree$\textrightarrow$50\degree$ towards $v_4$ is shorter than the rotation $20\degree$\textrightarrow$345\degree$ towards $v_7$. As such, the required sequence of actions to go from $v^1$ to $v^5$ in \citet{Chen2018Touchdown}'s environment is $[\act{forward}, \act{forward}, \act{forward}, \act{forward}]$. This is unpredictable and is not correctly aligned with "\textit{turn right at the intersection}" instructions.\footnote{In the Appendix we show more examples for 3-way, 4-way and 5-way intersections.} To alleviate this problem, \citet{schumann-riezler-2022-analyzing} explicitly feed the change of heading at each timestep as additional input to their model. This enables the agent to anticipate the unexpected rotation and to adapt to it. Because adding heading delta values to the text-based interface makes it convoluted and unnecessarily difficult for few-shot learning, we propose a more intuitive way to solve this ambiguity at intersections. We modify the state transition function $\phi$ such that the agent is not automatically rotated when moving $\act{forward}$. This means the agent's heading $\alpha_t$ is not automatically aligned with an outgoing edge. Instead, the direction is selected in relation to all outgoing edges. The agent at node $v^3$ in Figure~\ref{fig:intersection} has the nodes $v^6$, $v^7$ and $v^4$ in front. The forward direction is selected as the middle one of the three edges, the right direction as the right-most edge, and the left direction as the left-most edge. This means that executing the $\act{right}$ action at position $v^3$ will now rotate the agent towards node $v^4$ and allows to use the semantically correct sequence of actions for the depicted route. The proposed modification solves the issue of inconsistent action sequences at intersections and allows to use agents that are not specifically trained in this environment.
\subsection{Turning Around}
We additionally introduce the $\act{turn\_around}$ action which lets the agent reverse its direction: $s_{t+1} = (v, \alpha_t - 180\degree)$. In the unmodified environment, this is achieved using the $\act{left}$ or $\act{right}$ action on regular street segments. The new action is better aligned with natural language verbalizations of direction reversal and promotes intuitive communication with the environment.

\section{Navigation Task}
\label{sec:task}
The objective of the navigation task is to find the goal location by following the given navigation instructions. A navigation instance is defined by the initial state $s_1$, target node $\hat{v}_T$, gold path $(\hat{v}_1, \hat{v}_2..., \hat{v}_T)$ and navigation instructions text $n=(w_1, w_2, ..., w_N)$. The agent starts at $s_1$ and predicts the next action $a_1$ based on the navigation instructions and current observations. These are the panorama image and number of outgoing edges at the current position. The environment processes the action and puts the agent into a new state: $s_2 = \phi(a_1, s_1)$. This is repeated until the agent predicts $\act{stop}$ at the presumed goal location. If the agent stops within one neighboring node of the target node, the navigation objective is considered accomplished.

\begin{table}
\centering
%\ra{1.05}
\resizebox{.99\linewidth}{!}{
\begin{tabular}{l|l}
\toprule
\multicolumn{2}{c}{\textbf{Egocentric Spatial Reasoning}}\\ 
\midrule
1. & ... turn so the orange construction barrier is on your left ...\\
2. & ... a red truck in front of you ...\\
3. & ... a playground on the far right corner ahead ...\\
\midrule
\multicolumn{2}{c}{\textbf{Allocentric Spatial Reasoning}}\\ 
\midrule
4. & ... green metal pole with pink flowers on top  ...\\
5. & ... building with columns around the windows ...\\
6. & ... stop in between Chase and Dunkin' Donuts ... \\
\midrule
\multicolumn{2}{c}{\textbf{Temporal Reasoning}}\\ 
\midrule
7. & ... go straight until you see Chipotle and then ... \\
8. & ... once you passed the underpass ... \\
9. & ... stop when the park on your right ends ... \\
\midrule
\multicolumn{2}{c}{\textbf{Other}}\\ 
\midrule
10. & ... proceed straight through three more intersections ...\\
11. & ... you should see TD Bank on your left ...\\
12. & ... if you see Dory Oyster Bar, you have gone too far ...\\
\bottomrule
\end{tabular}
}
\caption{Reasoning skills the embodied LLM agent must possess in order to successfully complete the navigation task. Each with three example snippets from the navigation instructions.}
\label{tab:reasoning}
\end{table}

\subsection{Challenges}
\begin{figure*}[t]
    \centering
    \includegraphics[width=0.99\textwidth]{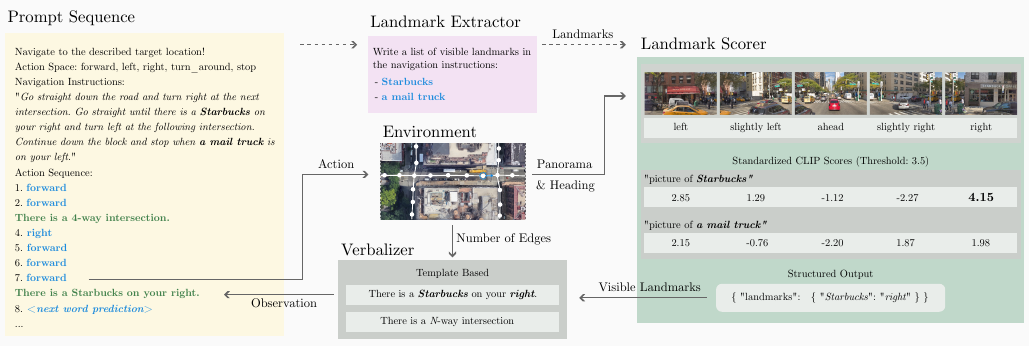}
    \caption{Overview of the proposed agent VELMA navigating in the Street View environment. The prompt sequence includes the task description, navigation instructions, and verbalized navigation trajectory up to the current timestep. The next action is decided by next word prediction utilizing an LLM and subsequently executed in the environment. This puts the agent into a new state, and the landmark scorer determines if an extracted landmark is visible in the current panorama view. The verbalizer takes this landmark information along with the information about a potential intersection and produces the current observations text. This text is then appended to the prompt sequence and again used to predict the next action. This process is repeated until the agent stops and the alleged target location.}
    \label{fig:workflow}
\end{figure*}
One main challenge to successfully follow the navigation instructions is to reliably detect landmarks in the panorama images along the route. The landmarks mentioned in the instructions are open-ended and can refer to any object or structure found in street scenes, including vegetation, building features, vehicle types, street signs, construction utilities, company logos and store names. The agent also needs to posses different types of reasoning, most importantly spatial reasoning to follow general directions, locate landmarks and evaluate stopping conditions. The agent also needs to understand the temporal aspect of the task and reason about the sequence of previous observations and actions. See Table~\ref{tab:reasoning} for example snippets from the navigation instructions.

\subsection{Datasets}
There are two datasets that provide navigation instructions for the environment described in Section~\ref{sec:env}: \textbf{Touchdown}~\cite{Chen2018Touchdown} and \textbf{Map2seq}~\cite{schumann-riezler-2021-map2seq}. Each dataset includes around 10k navigation instances, and we utilize them in the more challenging \textit{unseen} scenario introduced by \citet{schumann-riezler-2022-analyzing}. This means that generalization is crucial because the training routes are located in an area that is geographically separated from the area of development and test routes. The main difference between the two datasets is that Touchdown instructions were written by annotators who followed the route in Street View, while Map2seq instructions were written by annotators that saw a map of the route. The Map2seq navigation instructions were later validated to also be correct in Street View. Another difference is that the initial state in Map2seq orientates the agent towards the correct direction which leads to overall better task completion rates than for Touchdown instances.
\begin{figure*}[t]
  \centering
  \begin{subfigure}{0.49\textwidth}
  \includegraphics[width=\textwidth]{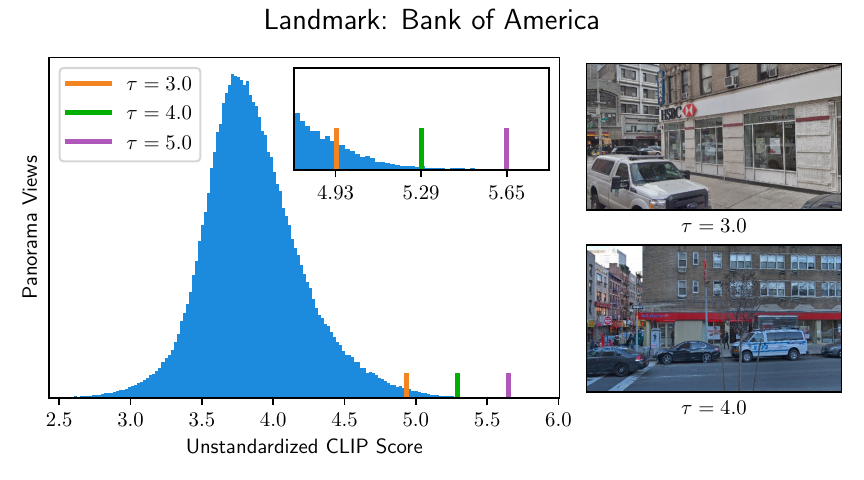}
  \end{subfigure}
  \begin{subfigure}{0.49\textwidth}
  \includegraphics[width=\textwidth]{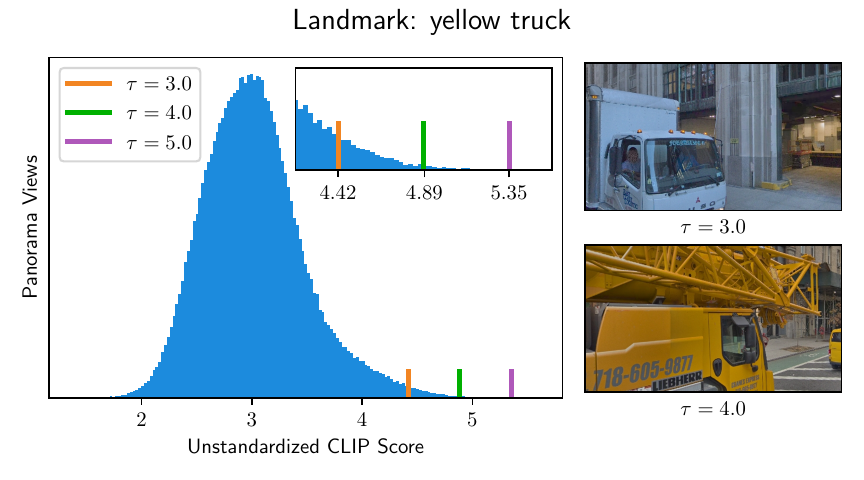}
  \end{subfigure}
  \caption{Distribution of CLIP scores between a landmark and panorama images in the training area. The CLIP score represents the semantic similarity of the panorama image and the text caption "picture of $[$landmark$]$". The distribution is used to standardize the score of the landmark and a novel panorama. The threshold $\tau$ is defined on the standardized score and used to determine the visibility of the landmark in the novel panorama image.}
  \label{fig:pano_threshold}
\end{figure*}

\section{LLM Agent}
In this section, we propose the urban VLN agent that uses an LLM to reason about the next action. To this end, we verbalize the navigation task, especially the environment observations. The workflow includes the extraction of landmarks that are mentioned in the instructions and determining their visibility in the current panorama image. The verbalizer then integrates the visible landmarks and street intersections into an observation text phrase $o_t$ at each step. The complete text prompt at timestep $t$ is composed as follows:
\begin{equation}
    x_t = [d^a, n, d^b, o_1, 1, a_1, o_2, 2, a_2, ..., o_t, t],
    \label{eq:prompt}
\end{equation}
where $[\;]$ denotes string concatenation, $d^a$ and $d^b$ are part of the task description and $n$ is the navigation instructions text. Punctuation and formatting are omitted in the notation for brevity. Figure~\ref{fig:workflow} shows a prompt sequence at $t=8$ on the left. This formulation of the navigation task enables the agent to predict the next action by next word prediction:
\begin{equation}
    a_{t} = \argminA_{w \in A} P_{LLM}(w|x_t),
    \label{eq:prediction}
\end{equation}
where $A$ are the literals of the five defined actions and $P_{LLM}$ is a black-box language model with no vision capabilities.

\subsection{Landmark Extractor}
Each navigation instructions text $n$ mentions multiple landmarks for visual guidance. In order to determine if a mentioned landmark is visible in the current panorama view, we first have to extract them from the instructions text. For this, we create a single prompt that includes five in-context examples of navigation instructions paired with a list of landmarks~(shown in the Appendix). It is used by the LLM to automatically generate the list of landmarks $(l_1, l_2, ..., l_L)$ mentioned in the given navigation instructions. The landmark extractor is depicted in the top middle of Figure~\ref{fig:workflow} and executed before the navigation starts. 

\subsection{Landmark Scorer}
At each step, the agent observes a panorama view $p^\alpha_v$, defined by its current position $v$ and heading direction $\alpha$. The view is an 800x460 sized image cut from the panorama with 60\degree~field-of-view. In order to determine if a landmark $l_i$ is visible in the view, we employ a CLIP model~\cite{Radford2021LearningTV} to embed the image and the caption: "\textit{picture of [$l_i$]}". The similarity score of the two embeddings determines the visibility of the landmark. Because the scores can be biased towards certain types of landmarks, we standardize them using all views $p^*_{train}$ of the \url{~}20k panorama images in the training area. Recall that we operate in the \textit{unseen scenario} where the training area and evaluation area are geographically separated. The standardized score of a landmark is:
\begin{equation}
\begin{gathered}
    z(l, p^\alpha_v) = \frac{\text{CLIP}(l, p^\alpha_v) - \mu(C_l)}{\sigma(C_l)}\\
    \text{where } C_l = \{\text{CLIP}(l,p^{\alpha'}_{v'}) \mid p^{\alpha'}_{v'} \in p^*_{train}\}.
\end{gathered}
\end{equation}
If the standardized score is larger than the threshold $\tau$, the landmark is classified as visible in the current view. The process does not require annotations and is completely unsupervised, allowing to score novel landmarks. The threshold is the only tunable parameter in the landmark scorer. Figure~\ref{fig:pano_threshold} shows the distribution of unstandardized CLIP scores and views at different threshold values for two example landmarks. While the views at $\tau=4.0$ both show the correct landmark, the view at $\tau=3.0$ for "Bank of America" shows an HSBC branch, and for "yellow truck" it shows a white truck. This suggests that the optimal threshold lies between the two values. As depicted on the right in Figure~\ref{fig:workflow}, the agent also evaluates views to the left and right of the current heading. Each panorama view direction $(p^{\alpha-90\degree}_v, p^{\alpha-45\degree}_v, p^{\alpha}_v, p^{\alpha+45\degree}_v, p^{\alpha+90\degree}_v)$ is associated with a string literal $m$ valued \textit{left}, \textit{slightly left}, \textit{ahead}, \textit{slightly right} or \textit{right}, respectively. A visible landmark $l_i$ and the corresponding direction literal $m_i$ are passed to the verbalizer. A full navigation trajectory includes around 200 image views~(40 steps and 5 view directions per step) and each landmark is typically visible in only one or two views.

\subsection{Verbalizer}
The verbalizer is a template-based component that produces environment observations in text form. There are two types of environment observations. First, there are street intersections that are detected based on the number of outgoing edges $N(v)$ at the current node $v$ in the navigation graph. If there are three or more outgoing edges at step $t$, the verbalizer encodes this information into the observation string~$o_t^e$: "\textit{There is a $[N(v)]$-way intersection}". Extracting this information directly from the navigation graph is akin to the junction type embedding used by the ORAR model~\cite{schumann-riezler-2022-analyzing} and is motivated by direction arrows displayed in the Street View GUI that human navigators used during data collection. The other type of observations are landmarks visible in the panorama view. The landmark name $l_i$ and direction literal $m_i$ are used to verbalize the observation $o_t^l$: "\textit{There is $[l_i]$ on your $[m_i]$}". The complete observation is $o_t=[o_t^e, o_t^l]$, where the respective string is empty if no intersection or landmark is detected. The observation is appended to the prompt in Equation~\ref{eq:prompt} and used by the agent to decide the next action. 

\section{Experiments}

\begin{table*}[t]
\centering
%\ra{1.05}
\resizebox{.99\linewidth}{!}{
\begin{tabular}{@{}lc@{}c@{}ccc@{}c@{}ccc@{}c@{}ccc@{}c@{}c@{}}
\toprule
&  \multicolumn{7}{c}{\textbf{Development Set}} & \phantom{} &\multicolumn{7}{c}{\textbf{Test Set}}\\ 
\cmidrule{2-8}  \cmidrule{10-16}
&  \multicolumn{3}{c}{\textbf{Touchdown}} & \phantom{} &\multicolumn{3}{c}{\textbf{Map2seq}} & \phantom{}  & \multicolumn{3}{c}{\textbf{Touchdown}} & \phantom{} &\multicolumn{3}{c}{\textbf{Map2seq}}\\ 
\cmidrule{2-4} \cmidrule{6-8} \cmidrule{10-12} \cmidrule{14-16}
    \textbf{Model} & \textbf{SPD\textdownarrow} & \textbf{KPA\textuparrow} & \textbf{TC\textuparrow} & \phantom{} & \textbf{SPD\textdownarrow} & \textbf{KPA\textuparrow} & \textbf{TC\textuparrow}& \phantom{}& \textbf{SPD\textdownarrow} & \textbf{KPA\textuparrow} & \textbf{TC\textuparrow} & \phantom{} & \textbf{SPD\textdownarrow} & \textbf{KPA\textuparrow} & \textbf{TC\textuparrow}\\
\toprule
\midrule
ORAR-ResNet      & 20.0\,\scriptsize{±0.7} & - & \enskip 15.4\,\scriptsize{±2.2} && 11.9\,\scriptsize{±0.4} & - & \enskip 27.6\,\scriptsize{±1.8} && 20.8\,\scriptsize{±0.6} & - & \enskip 14.9\,\scriptsize{±1.2} && 13.0\,\scriptsize{±0.3} & - & \enskip 30.3 \,\scriptsize{±1.8} \\
ORAR\textsuperscript{$\spadesuit$}-ResNet & 16.5\,\scriptsize{±0.1} & \enskip \textbf{64.0}\,\scriptsize{±0.2} & \enskip22.6\,\scriptsize{±0.6}  && 10.3\,\scriptsize{±0.4} & \enskip 74.4\,\scriptsize{±0.8} & \enskip 29.9\,\scriptsize{±1.7} && 17.4\,\scriptsize{±0.2} & \enskip 62.3\,\scriptsize{±0.1} & \enskip 19.1\,\scriptsize{±1.0}  && 10.9\,\scriptsize{±0.1} & \enskip74.7\,\scriptsize{±0.2} & \enskip32.5\,\scriptsize{±1.4} \\
ORAR\textsuperscript{$\spadesuit$}-OpenCLIP      & 17.5\,\scriptsize{±0.2} & \enskip 63.7\,\scriptsize{±1.0} & \enskip 21.5\,\scriptsize{±0.9} && 10.0\,\scriptsize{±0.2} & \enskip 75.3\,\scriptsize{±0.5} & \enskip 32.8\,\scriptsize{±1.5} && 17.0\,\scriptsize{±0.1} & \enskip  \textbf{63.4}\,\scriptsize{±0.4} & \enskip 20.0\,\scriptsize{±0.1}  && 10.5\,\scriptsize{±0.5} & \enskip 75.1\,\scriptsize{±0.7} & \enskip 34.0\,\scriptsize{±0.5} \\
\midrule
\midrule
&\multicolumn{15}{c}{\textbf{2-Shot In-Context Learning}}\\
\midrule
VELMA-Mixtral      & 28.4 \phantom{\,\scriptsize{±0.0}} & \enskip47.2 \phantom{\,\scriptsize{±0.0}} & \enskip6.5 \phantom{\,\scriptsize{±0.0}} && 21.1 \phantom{\,\scriptsize{±0.0}} & \enskip 56.8 \phantom{\,\scriptsize{±0.0}} & \enskip 8.0 \phantom{\,\scriptsize{±0.0}} && - & - & - && - & - & -\\
VELMA-GPT-3      & 22.2 \phantom{\,\scriptsize{±0.0}} & \enskip49.1 \phantom{\,\scriptsize{±0.0}} & \enskip6.8 \phantom{\,\scriptsize{±0.0}} && 19.1 \phantom{\,\scriptsize{±0.0}} & \enskip 58.1 \phantom{\,\scriptsize{±0.0}} & \enskip 9.2 \phantom{\,\scriptsize{±0.0}} && - & - & - && - & - & -\\
VELMA-GPT-4      & \underline{21.8} \phantom{\,\scriptsize{±0.0}} & \enskip\underline{56.1} \phantom{\,\scriptsize{±0.0}} & \enskip\underline{10.0} \phantom{\,\scriptsize{±0.0}} && \underline{12.8} \phantom{\,\scriptsize{±0.0}} & \enskip \underline{70.1} \phantom{\,\scriptsize{±0.0}} & \enskip \underline{23.1} \phantom{\,\scriptsize{±0.0}} && - & - & - && - & - & -\\
\midrule
\midrule
&\multicolumn{15}{c}{\textbf{LLM Finetuning, full training set}}\\
\midrule
VELMA-FT    & 18.3\,\scriptsize{±0.4} & \enskip62.0\,\scriptsize{±0.1} & \enskip23.4\,\scriptsize{±0.2} && 8.7\,\scriptsize{±0.0} & \enskip 78.7\,\scriptsize{±0.3} & \enskip 41.3\,\scriptsize{±0.9} && 18.2\,\scriptsize{±0.3} & \enskip 62.2\,\scriptsize{±0.2} & \enskip 23.5\,\scriptsize{±0.4} && 9.7\,\scriptsize{±0.3} & \enskip78.0\,\scriptsize{±0.1} & \enskip40.0\,\scriptsize{±1.0}\\
 VELMA-RBL     & \textbf{15.5}\,\scriptsize{±0.3} & \enskip63.6\,\scriptsize{±0.6} & \enskip\textbf{26.0}\,\scriptsize{±0.6} && \textbf{8.3}\,\scriptsize{±0.1} & \enskip \textbf{79.5}\,\scriptsize{±0.4} & \enskip \textbf{45.3}\,\scriptsize{±0.5} && \textbf{16.0}\,\scriptsize{±0.7} & \enskip 62.8\,\scriptsize{±1.3} & \enskip \textbf{26.4}\,\scriptsize{±1.7} && \textbf{8.3}\,\scriptsize{±0.2} & \enskip\textbf{79.6}\,\scriptsize{±0.4} & \enskip\textbf{47.5}\,\scriptsize{±0.7}\\
\bottomrule
\end{tabular}
}
\caption{Results for the urban VLN task on Touchdown and Map2seq in the \textit{unseen} scenario, meaning the training area is geographically separated from the area where development and test routes are located. ORAR-ResNet~\cite{schumann-riezler-2022-analyzing} is the previous best model and follows a seq-to-seq architecture that fuses text and image features during decoding. We retrained this model in our improved environment~(ORAR\textsuperscript{$\spadesuit$}-ResNet) and also with the same image feature extractor~(ORAR\textsuperscript{$\spadesuit$}-OpenCLIP) that we use in the landmark scorer. VELMA-GPT-3 and VELMA-GPT-4 models employ our proposed verbalization workflow and are prompted with two in-context examples. Due to cost and data leakage concerns, we evaluate the GPT models on the development sets only. VELMA-FT is LLaMa-7b finetuned on all training text sequences~(around 6k for each dataset). The VELMA-RBL finetuning process is described in Section~\ref{sec:rl}. All experiments are repeated three times with different random seeds (mean/std reported). \textbf{Bold} values are the nominal best results and \underline{underlined} are best few-shot results.}
\label{tab:results}
\end{table*}

We conducted experiments\footnote{Project page: \url{https://velma.schumann.pub/} and code: \url{https://github.com/raphael-sch/VELMA}} to evaluate the navigation performance of the proposed LLM agent in finetuning and in-context learning settings. We used \texttt{CLIP-ViT-bigG-14-laion2B-39B-b160k}~\cite{schuhmann2022laionb} as the CLIP model in the landmark scorer. We set the threshold $\tau=3.5$ for all experiments. The threshold was selected by inspecting the distribution of CLIP scores~(as in Figure~\ref{fig:pano_threshold}) for a handful of landmarks. On purpose, we did not systematically tune it in order to not violate the premise of few-shot learning.

\subsection{Landmark Extraction}
We ran the landmark extractor once for all instances using GPT-3~\cite{NEURIPS2020_1457c0d6} and used the same extracted landmarks in all experiments. On average, 2.7 landmarks were extracted from a navigation instructions text. Around 58\% of the landmarks in the test sets are \textit{novel}, i.e., they are not used in the training instances. In order to estimate the quality of the automatically extracted landmarks, we annotated 50 instances of each development set by hand. For Touchdown we calculated an F1-score of 96.3~(precision: 97.2, recall: 95.4) and the F1-score for Map2seq is 99.6~(precision: 100, recall: 99.3). This shows that GPT-3 reliably extracts landmarks from the instructions text and reusing them for all experiments is minimizing the inaccuracies introduced by this workflow step.

\subsection{Metrics and Baseline}
We use three metrics to measure navigation performance. The task completion~(TC) rate is a binary metric that measures whether the agent successfully stopped within one neighboring node of the target location. Shortest-path distance~(SPD) calculates the shortest path length between the stopping location and goal location~\cite{Chen2018Touchdown}. Key point accuracy~(KPA) measures the ratio of correct decisions at key points. Key points include the initial step, intersections along the gold route, and the target location.

For baselines, we use the current state-of-the-art agent model for urban VLN called ORAR~\cite{schumann-riezler-2022-analyzing}. The model employs a seq-to-seq architecture where the encoder LSTM reads the navigation instructions text, and the multi-layer decoder LSTM receives image feature vectors of the current panorama view as additional input at each action decoding step. The ORAR model is a very strong baseline beating more sophisticated models like the VLN Transformer~\cite{zhu-etal-2021-multimodal}. Because the environment modifications introduced in Section~\ref{sec:env} spare the agents from learning specific irregularities, we additionally retrain ORAR in the improved environment for a fair comparison.

\subsection{Few-Shot Learning Results}
\begin{figure}[!ht]
    \centering
    \includegraphics[width=0.48\textwidth]{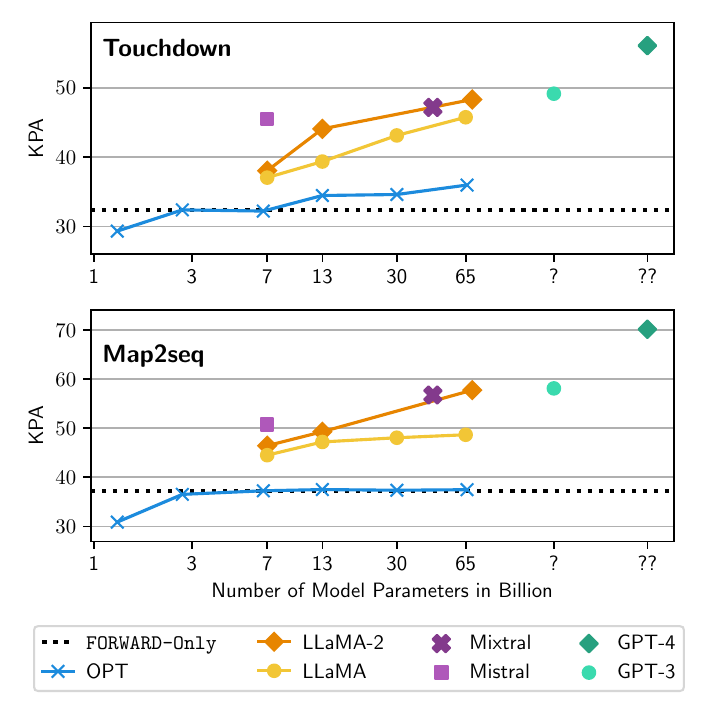}
    \caption{Key point accuracy (KPA) for 2-shot in-context learning of large language models with increasing parameter count. The $\act{forward}$-Only baseline predicts walking forward until the average trajectory length is reached and performs better than predicting random directions.}
    \label{fig:llm_inference}
\end{figure}
The proposed text-only interface allows us to use large language models as reasoners without updating their weights or fusing image representations. The prompt consists of a short task description and two in-context examples~(2-shot). The examples are full text sequences along the gold route for randomly selected navigation instances in the training set. The two plots in Figure~\ref{fig:llm_inference} show that performance scales with parameter count and varies across model families. The $\act{forward}$-Only baseline reveals that OPT~\cite{zhang2022opt} can barely compete with a basic heuristic, even at a model size of 65~billion parameters. LLaMa~\cite{touvron2023llama} and especially LLaMa-2~\cite{touvron2023llama2} show promising navigation skills reaching 48.3 and 57.7 key point accuracy~(KPA) on Touchdown and Map2seq, respectively. However, this KPA score only translates to task completion~(TC) rates of 2.1 and 3.2, revealing that the model is not able to consistently predict correct actions throughout the whole navigation trajectory. Mistral-7b performs on par with a LLaMA-2 model twice its size, but also fails to score task completion rates significantly higher than 3. The only few-shot LLMs that achieve substantial TC rates are GPT-3, GPT-4~\cite{OpenAI2023GPT4TR} and Mixtral~\cite{mistralAI2023mixtral}. As listed in Table~\ref{tab:results}, VELMA-GPT-4 achieves the best results for the 2-shot setting. It reaches 44\% and 77\% of the TC rate reported for the previous state-of-the-art model ORAR\textsuperscript{$\spadesuit$}-ResNet which is a seq-to-seq model that has direct access to image features and was trained on the full training set. In contrast, the LLMs in our work act as a blind agent that solely relies on observation descriptions produced by the verbalizer. This is remarkable because LLMs are not explicitly trained to experience embodiment in a visual environment. This is emergent behavior unearthed by verbalizing the VLN task. We also observe that GPT-4 invokes the $\act{turn\_around}$ action in useful ways, e.g. to return a few steps when it notices that it went past the described goal location. This emphasizes the effectiveness of intuitive communication with the environment.

\subsection{Finetuning Results}
\label{sec:finetune}
To further explore the capabilities of the proposed LLM agent, we finetune LLaMa-7b on all training instances of the respective dataset, denoted by VELMA-FT in Table~\ref{tab:results}. Each training instance is the full text sequence that is produced by following the gold path. The visibility of landmarks is determined by the landmark scorer during training because gold annotations are not available. There are 6,770 training instances for Touchdown and 5,737 for Map2seq. We finetune for 20 epochs using LoRA~\cite{hu2022lora} to adapt query, key and value projections of the attention layer as well as input and output projection of each transformer layer. The best model is selected by task completion on the development set. The resulting agent outperforms the previous state-of-the-art model ORAR\textsuperscript{*} by 10\% and 16\% relative TC rate. Comparing ORAR\textsuperscript{*} which fuses image features at the vector level to VELMA-FT which finetunes on verbalizations of observations, shows that the text-based environment observations are less prone to overfitting.

\subsubsection{Response-Based Learning}

\begin{table}[t]
\centering
%\ra{1.05}
\resizebox{.99\linewidth}{!}{
\begin{tabular}{llccccc}
\toprule
& \phantom{}  & \multicolumn{2}{c}{\textbf{Touchdown}} &\phantom{}& \multicolumn{2}{c}{\textbf{Map2seq}}\\ 
\cmidrule{3-4} \cmidrule{6-7}
    \textbf{Image Model} & \phantom{}& \textbf{SPD\textdownarrow} & \textbf{TC\textuparrow} &\phantom{}& \textbf{SPD\textdownarrow} & \textbf{TC\textuparrow}\\
\toprule
no image && 26.7 \scriptsize{±0.4} & 15.0 \scriptsize{±0.6} && 9.1 \scriptsize{±0.2} & 37.8 \scriptsize{±1.0}\\
CLIP && 20.2 \scriptsize{±0.3} & 20.8 \scriptsize{±0.5} && 8.8 \scriptsize{±0.5} & 39.2 \scriptsize{±0.5}\\
OpenCLIP && \textbf{18.3} \scriptsize{±0.4} & \textbf{23.4} \scriptsize{±0.2} && \textbf{8.7} \scriptsize{±0.0} & \textbf{41.3} \scriptsize{±0.9}\\
\bottomrule
\end{tabular}
}
\caption{Vision ablation on the development set. We finetune a separate LLaMa-7b model for each ablation. CLIP refers to \texttt{clip-vit-large-patch14} \cite{Radford2021LearningTV}. The OpenCLIP image model refers to \texttt{CLIP-ViT-bigG-14-laion2B-39B-b160k}~\cite{schuhmann2022laionb}.}
\label{tab:results_image}
\end{table}

\label{sec:rl}
A navigation task is successfully completed if the agent stops at either the goal location or an adjacent neighboring node. Training the agent with teacher-forcing to exactly follow the gold route penalizes the agent for stopping one step short or one step past the target node, despite accomplishing the navigation objective. Furthermore, the agent can not learn to recover from incorrect decisions during inference. We thus train the agent to directly optimize the TC metric while also feeding it its own actions during training, called VELMA-RBL in Table~\ref{tab:results}. The procedure for VELMA-RBL is inspired by response-based learning~\cite{clarke-etal-2010-driving} and imitation learning~\cite{RossETAL:11} and is outlined in Algorithm~\ref{alg:rl}. The loss for an instance at training step $j$ is either computed by teacher forcing the gold action sequence $\hat{\mathbf{a}}$, or by student forcing, determined by a mixing parameter $\lambda$. In student forcing, the actions decoded by the current model weights $\theta_j$ are executed instead of the gold actions. If this trajectory ends within one neighboring node of the target location, the predicted action sequence $\mathbf{a}_j$ is considered correct and used as the reference to train the agent. If the agent stops at the wrong location, an oracle path is computed to provide the optimal counterfactual action at each step in the trajectory. In our case, the oracle's optimal next action is computed as the shortest path to the goal location. We set $\lambda=0.5$ to collect training losses in a batch evenly from both training strategies. Manually inspecting trajectories produced by the trained agent, we found improvements of following instructions that have stopping criteria like "\textit{Stop a few steps before Y.}" or "\textit{Stop at X. If you see Y you have gone too far.}". In both cases, the agent learned to walk past the uncertain stopping location and to invoke the $\act{turn\_around}$ action in order to walk back once landmark Y appeared. The described training procedure leads to a significant increase of task completion rate by 2.9 and 7.5 for Touchdown and Map2seq, respectively. Overall, our contributions in this work amount to a relative increase of task completion by 77\% and 57\% over the previously reported state-of-the-art for urban VLN on the Touchdown and Map2seq datasets.
\begin{algorithm}[t]
\caption{RBL Optimization of Task Completion}\label{alg:cap}
\begin{algorithmic}
\Require mixing ratio $\lambda$, training step $j$, model~weights $\theta_j$, gold action sequence $\hat{\mathbf{a}}$, prompt $x_1$
    \If{$random(0, 1) < \lambda$}
    \State $\mathbf{a}_{\theta_j} = StudentForcing(\theta_j, x_1)$
    \State $\mathbf{a}_j = \argminA \mathbf{a}_{\theta_j}$
    \If{$TaskCompletion(\mathbf{a}_{j}) =1$}
    \State $loss_{j} = \mathcal{L}_{CE}(\mathbf{a}_{\theta_j}, \mathbf{a}_{j})$
    \Else
    \State ${\mathbf{a}}^\ast_{j} = Oracle_{stepwise}(\mathbf{a}_{j})$
    \State $loss_{j} = \mathcal{L}_{CE}(\mathbf{a}_{\theta_j}, {\mathbf{a}}^\ast_{j})$
    \EndIf
    \Else
    \State $\mathbf{a}_{\theta_j} = TeacherForcing(\theta_j, x_1, \hat{\mathbf{a}})$
    \State $loss_j = \mathcal{L}_{CE}(\mathbf{a}_{\theta_j}, \hat{\mathbf{a}})$
    \EndIf
\end{algorithmic}
\label{alg:rl}
\end{algorithm}

\subsection{Image Ablation}
In this section, we ablate the image model used by the landmark scorer. We finetune a LLaMa-7b model according to Section~\ref{sec:finetune} and use CLIP~\cite{Radford2021LearningTV}, OpenCLIP~\cite{schuhmann2022laionb} or no image model in the landmark scorer. The latter case means that no landmark observation is passed to the prompt sequence. The results in Table~\ref{tab:results_image} show that OpenCLIP is better suited for detecting landmarks in our navigation task than the original CLIP model. This is in line with better ImageNet results reported by the OpenCLIP authors and suggests that the agent can directly benefit from further improvements of CLIP models. Appending no landmarks to the prompt sequence further degrades performance, especially on Touchdown.

\section{Conclusion}
We introduced VELMA, an agent for urban vision and language navigation, which utilizes a large language model to infer its next action. The LLM is continuously queried with a text prompt that verbalizes the task description, navigation instructions, visual observations, and past trajectory of the agent. In order to include observed landmarks in the prompt, we propose an unsupervised pipeline that extracts landmarks from the instructions and determines their visibility in the current panorama view based on thresholded CLIP scores. We evaluate the embodied LLM agent in a modified version of the commonly used Touchdown environment based on Street View. One proposed modification is fixing a problem at intersections that led to incorrect alignments of action sequences, and another modification adds the $\act{turn\_around}$ action which provides a more intuitive way to communicate with the environment. The proposed agent achieves strong few-shot in-context learning results of 10 and 23 task completion rates for Touchdown and Map2seq, respectively, and yields new state-of-the-art results of 26 and 47 task completion rates when finetuned on the full training set. The finetuning results show that verbalization is not an inherent limitation for this task and in-context learning with better base models or improved prompting techniques could outperform our reported few shot results.

\section*{Acknowledgments}
The research reported in this paper was supported by a Google Focused Research Award.

\bibliography{aaai24}

\begin{thebibliography}{41}
\providecommand{\natexlab}[1]{#1}

\bibitem[{Anderson et~al.(2018)Anderson, Wu, Teney, Bruce, Johnson, S{\"u}nderhauf, Reid, Gould, and Van Den~Hengel}]{anderson2018vision}
Anderson, P.; Wu, Q.; Teney, D.; Bruce, J.; Johnson, M.; S{\"u}nderhauf, N.; Reid, I.; Gould, S.; and Van Den~Hengel, A. 2018.
\newblock Vision-and-language navigation: Interpreting visually-grounded navigation instructions in real environments.
\newblock In \emph{Proceedings of the IEEE conference on computer vision and pattern recognition}, 3674--3683.

\bibitem[{Armitage, Impett, and Sennrich(2023)}]{armitage2023priority}
Armitage, J.; Impett, L.; and Sennrich, R. 2023.
\newblock A Priority Map for Vision-and-Language Navigation with Trajectory Plans and Feature-Location Cues.
\newblock In \emph{Proceedings of the IEEE/CVF Winter Conference on Applications of Computer Vision}, 1094--1103.

\bibitem[{Brown et~al.(2020)Brown, Mann, Ryder, Subbiah, Kaplan, Dhariwal, Neelakantan, Shyam, Sastry, Askell, Agarwal, Herbert-Voss, Krueger, Henighan, Child, Ramesh, Ziegler, Wu, Winter, Hesse, Chen, Sigler, Litwin, Gray, Chess, Clark, Berner, McCandlish, Radford, Sutskever, and Amodei}]{NEURIPS2020_1457c0d6}
Brown, T.; Mann, B.; Ryder, N.; Subbiah, M.; Kaplan, J.~D.; Dhariwal, P.; Neelakantan, A.; Shyam, P.; Sastry, G.; Askell, A.; Agarwal, S.; Herbert-Voss, A.; Krueger, G.; Henighan, T.; Child, R.; Ramesh, A.; Ziegler, D.; Wu, J.; Winter, C.; Hesse, C.; Chen, M.; Sigler, E.; Litwin, M.; Gray, S.; Chess, B.; Clark, J.; Berner, C.; McCandlish, S.; Radford, A.; Sutskever, I.; and Amodei, D. 2020.
\newblock Language Models are Few-Shot Learners.
\newblock In Larochelle, H.; Ranzato, M.; Hadsell, R.; Balcan, M.; and Lin, H., eds., \emph{Advances in Neural Information Processing Systems}, volume~33, 1877--1901. Curran Associates, Inc.

\bibitem[{Chen et~al.(2019)Chen, Suhr, Misra, Snavely, and Artzi}]{Chen2018Touchdown}
Chen, H.; Suhr, A.; Misra, D.; Snavely, N.; and Artzi, Y. 2019.
\newblock TOUCHDOWN: Natural Language Navigation and Spatial Reasoning in Visual Street Environments.
\newblock In \emph{Proceedings of the IEEE/CVF Conference on Computer Vision and Pattern Recognition (CVPR)}. Long Beach, California.

\bibitem[{Chen et~al.(2021)Chen, Guhur, Schmid, and Laptev}]{chen2021history}
Chen, S.; Guhur, P.-L.; Schmid, C.; and Laptev, I. 2021.
\newblock History aware multimodal transformer for vision-and-language navigation.
\newblock \emph{Advances in neural information processing systems}, 34: 5834--5847.

\bibitem[{Clarke et~al.(2010)Clarke, Goldwasser, Chang, and Roth}]{clarke-etal-2010-driving}
Clarke, J.; Goldwasser, D.; Chang, M.-W.; and Roth, D. 2010.
\newblock Driving Semantic Parsing from the World{'}s Response.
\newblock In \emph{Proceedings of the Fourteenth Conference on Computational Natural Language Learning}, 18--27. Uppsala, Sweden: Association for Computational Linguistics.

\bibitem[{Dorbala et~al.(2022)Dorbala, Sigurdsson, Piramuthu, Thomason, and Sukhatme}]{Dorbala2022}
Dorbala, V.~S.; Sigurdsson, G.; Piramuthu, R.; Thomason, J.; and Sukhatme, G.~S. 2022.
\newblock Clip-nav: Using clip for zero-shot vision-and-language navigation.
\newblock In \emph{CoRL 2022 Workshop on Language and Robot Learning}.

\bibitem[{Fried et~al.(2018)Fried, Hu, Cirik, Rohrbach, Andreas, Morency, Berg-Kirkpatrick, Saenko, Klein, and Darrell}]{fried2018speaker}
Fried, D.; Hu, R.; Cirik, V.; Rohrbach, A.; Andreas, J.; Morency, L.-P.; Berg-Kirkpatrick, T.; Saenko, K.; Klein, D.; and Darrell, T. 2018.
\newblock Speaker-Follower Models for Vision-and-Language Navigation.
\newblock In \emph{Neural Information Processing Systems (NeurIPS)}.

\bibitem[{Fu et~al.(2020)Fu, Wang, Peterson, Grafton, Eckstein, and Wang}]{fu2020counterfactual}
Fu, T.-J.; Wang, X.~E.; Peterson, M.~F.; Grafton, S.~T.; Eckstein, M.~P.; and Wang, W.~Y. 2020.
\newblock Counterfactual vision-and-language navigation via adversarial path sampler.
\newblock In \emph{Computer Vision--ECCV 2020: 16th European Conference, Glasgow, UK, August 23--28, 2020, Proceedings, Part VI 16}, 71--86. Springer.

\bibitem[{Hermann et~al.(2020)Hermann, Malinowski, Mirowski, Banki-Horvath, Anderson, and Hadsell}]{directions_in_streetview}
Hermann, K.~M.; Malinowski, M.; Mirowski, P.; Banki-Horvath, A.; Anderson, K.; and Hadsell, R. 2020.
\newblock Learning to Follow Directions in Street View.
\newblock In \emph{Proceedings of the AAAI Conference on Artificial Intelligence (AAAI)}. New York, New York.

\bibitem[{Hong et~al.(2021)Hong, Wu, Qi, Rodriguez-Opazo, and Gould}]{hong2021vln}
Hong, Y.; Wu, Q.; Qi, Y.; Rodriguez-Opazo, C.; and Gould, S. 2021.
\newblock Vln bert: A recurrent vision-and-language bert for navigation.
\newblock In \emph{Proceedings of the IEEE/CVF conference on Computer Vision and Pattern Recognition}, 1643--1653.

\bibitem[{Hu et~al.(2022)Hu, yelong shen, Wallis, Allen-Zhu, Li, Wang, Wang, and Chen}]{hu2022lora}
Hu, E.~J.; yelong shen; Wallis, P.; Allen-Zhu, Z.; Li, Y.; Wang, S.; Wang, L.; and Chen, W. 2022.
\newblock Lo{RA}: Low-Rank Adaptation of Large Language Models.
\newblock In \emph{International Conference on Learning Representations}.

\bibitem[{Khandelwal et~al.(2022)Khandelwal, Weihs, Mottaghi, and Kembhavi}]{Khandelwal_2022_CVPR}
Khandelwal, A.; Weihs, L.; Mottaghi, R.; and Kembhavi, A. 2022.
\newblock Simple but Effective: CLIP Embeddings for Embodied AI.
\newblock In \emph{Proceedings of the IEEE/CVF Conference on Computer Vision and Pattern Recognition (CVPR)}, 14829--14838.

\bibitem[{Kingma and Ba(2015)}]{adam}
Kingma, D.~P.; and Ba, J. 2015.
\newblock Adam: A Method for Stochastic Optimization.
\newblock In \emph{Proceedings of the International Conference on Learning Representations (ICLR)}. San Diego, California.

\bibitem[{Ku et~al.(2020)Ku, Anderson, Patel, Ie, and Baldridge}]{ku2020room}
Ku, A.; Anderson, P.; Patel, R.; Ie, E.; and Baldridge, J. 2020.
\newblock Room-Across-Room: Multilingual Vision-and-Language Navigation with Dense Spatiotemporal Grounding.
\newblock In \emph{Proceedings of the 2020 Conference on Empirical Methods in Natural Language Processing (EMNLP)}, 4392--4412.

\bibitem[{Li, Tan, and Bansal(2022)}]{li2022envedit}
Li, J.; Tan, H.; and Bansal, M. 2022.
\newblock Envedit: Environment editing for vision-and-language navigation.
\newblock In \emph{Proceedings of the IEEE/CVF Conference on Computer Vision and Pattern Recognition}, 15407--15417.

\bibitem[{Mehta et~al.(2020)Mehta, Artzi, Baldridge, Ie, and Mirowski}]{Mehta2020Retouchdown}
Mehta, H.; Artzi, Y.; Baldridge, J.; Ie, E.; and Mirowski, P. 2020.
\newblock Retouchdown: Releasing Touchdown on {S}treet{L}earn as a Public Resource for Language Grounding Tasks in Street View.
\newblock In \emph{Proceedings of the Third International Workshop on Spatial Language Understanding (SpLU)}. Online.

\bibitem[{{Mistral AI Team}(2023)}]{mistralAI2023mixtral}
{Mistral AI Team}. 2023.
\newblock Mixtral of Experts: A High Quality Sparse Mixture-of-Experts.
\newblock \emph{Mistral AI Blog}.
\newblock Accessed: December 18, 2023.

\bibitem[{OpenAI(2023)}]{OpenAI2023GPT4TR}
OpenAI. 2023.
\newblock GPT-4 Technical Report.
\newblock \emph{ArXiv}, abs/2303.08774.

\bibitem[{Qi et~al.(2020)Qi, Wu, Anderson, Wang, Wang, Shen, and Hengel}]{qi2020reverie}
Qi, Y.; Wu, Q.; Anderson, P.; Wang, X.; Wang, W.~Y.; Shen, C.; and Hengel, A. v.~d. 2020.
\newblock Reverie: Remote embodied visual referring expression in real indoor environments.
\newblock In \emph{Proceedings of the IEEE/CVF Conference on Computer Vision and Pattern Recognition}, 9982--9991.

\bibitem[{Radford et~al.(2021)Radford, Kim, Hallacy, Ramesh, Goh, Agarwal, Sastry, Askell, Mishkin, Clark, Krueger, and Sutskever}]{Radford2021LearningTV}
Radford, A.; Kim, J.~W.; Hallacy, C.; Ramesh, A.; Goh, G.; Agarwal, S.; Sastry, G.; Askell, A.; Mishkin, P.; Clark, J.; Krueger, G.; and Sutskever, I. 2021.
\newblock Learning Transferable Visual Models From Natural Language Supervision.
\newblock In \emph{ICML}.

\bibitem[{Ramakrishnan et~al.(2021)Ramakrishnan, Gokaslan, Wijmans, Maksymets, Clegg, Turner, Undersander, Galuba, Westbury, Chang, Savva, Zhao, and Batra}]{ramakrishnan2021habitatmatterport}
Ramakrishnan, S.~K.; Gokaslan, A.; Wijmans, E.; Maksymets, O.; Clegg, A.; Turner, J.~M.; Undersander, E.; Galuba, W.; Westbury, A.; Chang, A.~X.; Savva, M.; Zhao, Y.; and Batra, D. 2021.
\newblock Habitat-Matterport 3D Dataset ({HM}3D): 1000 Large-scale 3D Environments for Embodied {AI}.
\newblock In \emph{Thirty-fifth Conference on Neural Information Processing Systems Datasets and Benchmarks Track (Round 2)}.

\bibitem[{Ross, Gordon, and Bagnell(2011)}]{RossETAL:11}
Ross, S.; Gordon, G.~J.; and Bagnell, J.~A. 2011.
\newblock A Reduction of Imitation Learning and Structured Prediction to No-Regret Online Learning.
\newblock In \emph{Proceedings of the 14th International Conference on Artificial Intelligence and Statistics {(AISTATS)}}. Fort Lauderdale, {FL, USA}.

\bibitem[{Schuhmann et~al.(2022)Schuhmann, Beaumont, Vencu, Gordon, Wightman, Cherti, Coombes, Katta, Mullis, Wortsman, Schramowski, Kundurthy, Crowson, Schmidt, Kaczmarczyk, and Jitsev}]{schuhmann2022laionb}
Schuhmann, C.; Beaumont, R.; Vencu, R.; Gordon, C.~W.; Wightman, R.; Cherti, M.; Coombes, T.; Katta, A.; Mullis, C.; Wortsman, M.; Schramowski, P.; Kundurthy, S.~R.; Crowson, K.; Schmidt, L.; Kaczmarczyk, R.; and Jitsev, J. 2022.
\newblock {LAION}-5B: An open large-scale dataset for training next generation image-text models.
\newblock In \emph{Thirty-sixth Conference on Neural Information Processing Systems Datasets and Benchmarks Track}.

\bibitem[{Schumann and Riezler(2021)}]{schumann-riezler-2021-map2seq}
Schumann, R.; and Riezler, S. 2021.
\newblock Generating Landmark Navigation Instructions from Maps as a Graph-to-Text Problem.
\newblock In \emph{Proceedings of the 59th Annual Meeting of the Association for Computational Linguistics and the 11th International Joint Conference on Natural Language Processing (Volume 1: Long Papers)}, 489--502. Online: Association for Computational Linguistics.

\bibitem[{Schumann and Riezler(2022)}]{schumann-riezler-2022-analyzing}
Schumann, R.; and Riezler, S. 2022.
\newblock Analyzing Generalization of Vision and Language Navigation to Unseen Outdoor Areas.
\newblock In \emph{Proceedings of the 60th Annual Meeting of the Association for Computational Linguistics (Volume 1: Long Papers)}, 7519--7532. Dublin, Ireland: Association for Computational Linguistics.

\bibitem[{Shah et~al.(2022)Shah, Osinski, Ichter, and Levine}]{shah2022lmnav}
Shah, D.; Osinski, B.; Ichter, B.; and Levine, S. 2022.
\newblock LM-Nav: Robotic Navigation with Large Pre-Trained Models of Language, Vision, and Action.
\newblock arXiv:2207.04429.

\bibitem[{Shridhar et~al.(2020)Shridhar, Thomason, Gordon, Bisk, Han, Mottaghi, Zettlemoyer, and Fox}]{ALFRED20}
Shridhar, M.; Thomason, J.; Gordon, D.; Bisk, Y.; Han, W.; Mottaghi, R.; Zettlemoyer, L.; and Fox, D. 2020.
\newblock {ALFRED: A Benchmark for Interpreting Grounded Instructions for Everyday Tasks}.
\newblock In \emph{The IEEE Conference on Computer Vision and Pattern Recognition (CVPR)}.

\bibitem[{Sun et~al.(2023)Sun, Qiu, Aoki, and Kataoka}]{s23136028}
Sun, Y.; Qiu, Y.; Aoki, Y.; and Kataoka, H. 2023.
\newblock Outdoor Vision-and-Language Navigation Needs Object-Level Alignment.
\newblock \emph{Sensors}, 23(13).

\bibitem[{Tan, Yu, and Bansal(2019)}]{tan-etal-2019-learning}
Tan, H.; Yu, L.; and Bansal, M. 2019.
\newblock Learning to Navigate Unseen Environments: Back Translation with Environmental Dropout.
\newblock In \emph{Proceedings of the 2019 Conference of the North {A}merican Chapter of the Association for Computational Linguistics: Human Language Technologies, Volume 1 (Long and Short Papers)}, 2610--2621. Minneapolis, Minnesota: Association for Computational Linguistics.

\bibitem[{Touvron et~al.(2023{\natexlab{a}})Touvron, Lavril, Izacard, Martinet, Lachaux, Lacroix, Rozière, Goyal, Hambro, Azhar, Rodriguez, Joulin, Grave, and Lample}]{touvron2023llama}
Touvron, H.; Lavril, T.; Izacard, G.; Martinet, X.; Lachaux, M.-A.; Lacroix, T.; Rozière, B.; Goyal, N.; Hambro, E.; Azhar, F.; Rodriguez, A.; Joulin, A.; Grave, E.; and Lample, G. 2023{\natexlab{a}}.
\newblock LLaMA: Open and Efficient Foundation Language Models.
\newblock arXiv:2302.13971.

\bibitem[{Touvron et~al.(2023{\natexlab{b}})Touvron, Martin, Stone, Albert, Almahairi, Babaei, Bashlykov, Batra, Bhargava, Bhosale, Bikel, Blecher, Ferrer, Chen, Cucurull, Esiobu, Fernandes, Fu, Fu, Fuller, Gao, Goswami, Goyal, Hartshorn, Hosseini, Hou, Inan, Kardas, Kerkez, Khabsa, Kloumann, Korenev, Koura, Lachaux, Lavril, Lee, Liskovich, Lu, Mao, Martinet, Mihaylov, Mishra, Molybog, Nie, Poulton, Reizenstein, Rungta, Saladi, Schelten, Silva, Smith, Subramanian, Tan, Tang, Taylor, Williams, Kuan, Xu, Yan, Zarov, Zhang, Fan, Kambadur, Narang, Rodriguez, Stojnic, Edunov, and Scialom}]{touvron2023llama2}
Touvron, H.; Martin, L.; Stone, K.; Albert, P.; Almahairi, A.; Babaei, Y.; Bashlykov, N.; Batra, S.; Bhargava, P.; Bhosale, S.; Bikel, D.; Blecher, L.; Ferrer, C.~C.; Chen, M.; Cucurull, G.; Esiobu, D.; Fernandes, J.; Fu, J.; Fu, W.; Fuller, B.; Gao, C.; Goswami, V.; Goyal, N.; Hartshorn, A.; Hosseini, S.; Hou, R.; Inan, H.; Kardas, M.; Kerkez, V.; Khabsa, M.; Kloumann, I.; Korenev, A.; Koura, P.~S.; Lachaux, M.-A.; Lavril, T.; Lee, J.; Liskovich, D.; Lu, Y.; Mao, Y.; Martinet, X.; Mihaylov, T.; Mishra, P.; Molybog, I.; Nie, Y.; Poulton, A.; Reizenstein, J.; Rungta, R.; Saladi, K.; Schelten, A.; Silva, R.; Smith, E.~M.; Subramanian, R.; Tan, X.~E.; Tang, B.; Taylor, R.; Williams, A.; Kuan, J.~X.; Xu, P.; Yan, Z.; Zarov, I.; Zhang, Y.; Fan, A.; Kambadur, M.; Narang, S.; Rodriguez, A.; Stojnic, R.; Edunov, S.; and Scialom, T. 2023{\natexlab{b}}.
\newblock Llama 2: Open Foundation and Fine-Tuned Chat Models.
\newblock arXiv:2307.09288.

\bibitem[{Wang et~al.(2023)Wang, Xie, Jiang, Mandlekar, Xiao, Zhu, Fan, and Anandkumar}]{wang2023voyager}
Wang, G.; Xie, Y.; Jiang, Y.; Mandlekar, A.; Xiao, C.; Zhu, Y.; Fan, L.; and Anandkumar, A. 2023.
\newblock Voyager: An Open-Ended Embodied Agent with Large Language Models.
\newblock \emph{arXiv preprint arXiv: Arxiv-2305.16291}.

\bibitem[{Wang et~al.(2019)Wang, Huang, Celikyilmaz, Gao, Shen, Wang, Wang, and Zhang}]{Wang_2019_CVPR}
Wang, X.; Huang, Q.; Celikyilmaz, A.; Gao, J.; Shen, D.; Wang, Y.-F.; Wang, W.~Y.; and Zhang, L. 2019.
\newblock Reinforced Cross-Modal Matching and Self-Supervised Imitation Learning for Vision-Language Navigation.
\newblock In \emph{Proceedings of the IEEE/CVF Conference on Computer Vision and Pattern Recognition (CVPR)}.

\bibitem[{Xiang, Wang, and Wang(2020)}]{Xiang2020LearningNavigation}
Xiang, J.; Wang, X.; and Wang, W.~Y. 2020.
\newblock Learning to Stop: A Simple yet Effective Approach to Urban Vision-Language Navigation.
\newblock In \emph{Findings of the Association for Computational Linguistics (ACL Findings)}. Online.

\bibitem[{Zhang et~al.(2022)Zhang, Roller, Goyal, Artetxe, Chen, Chen, Dewan, Diab, Li, Lin et~al.}]{zhang2022opt}
Zhang, S.; Roller, S.; Goyal, N.; Artetxe, M.; Chen, M.; Chen, S.; Dewan, C.; Diab, M.; Li, X.; Lin, X.~V.; et~al. 2022.
\newblock Opt: Open pre-trained transformer language models.
\newblock \emph{arXiv preprint arXiv:2205.01068}.

\bibitem[{Zhong et~al.(2021)Zhong, Hanjie, Wang, Narasimhan, and Zettlemoyer}]{NEURIPS2021_b3e3e393}
Zhong, V.; Hanjie, A.~W.; Wang, S.; Narasimhan, K.; and Zettlemoyer, L. 2021.
\newblock SILG: The Multi-domain Symbolic Interactive Language Grounding Benchmark.
\newblock In Ranzato, M.; Beygelzimer, A.; Dauphin, Y.; Liang, P.; and Vaughan, J.~W., eds., \emph{Advances in Neural Information Processing Systems}, volume~34, 21505--21519. Curran Associates, Inc.

\bibitem[{Zhou, Hong, and Wu(2023)}]{zhou2023navgpt}
Zhou, G.; Hong, Y.; and Wu, Q. 2023.
\newblock NavGPT: Explicit Reasoning in Vision-and-Language Navigation with Large Language Models.
\newblock arXiv:2305.16986.

\bibitem[{Zhou et~al.(2023)Zhou, Zheng, Pryor, Shen, Jin, Getoor, and Wang}]{zhou2023esc}
Zhou, K.; Zheng, K.; Pryor, C.; Shen, Y.; Jin, H.; Getoor, L.; and Wang, X.~E. 2023.
\newblock ESC: Exploration with Soft Commonsense Constraints for Zero-shot Object Navigation.
\newblock \emph{arXiv preprint arXiv:2301.13166}.

\bibitem[{Zhu et~al.(2020)Zhu, Hu, Chen, Deng, Jain, Ie, and Sha}]{zhu2020babywalk}
Zhu, W.; Hu, H.; Chen, J.; Deng, Z.; Jain, V.; Ie, E.; and Sha, F. 2020.
\newblock BabyWalk: Going Farther in Vision-and-Language Navigation by Taking Baby Steps.
\newblock In \emph{Proceedings of the 58th Annual Meeting of the Association for Computational Linguistics}, 2539--2556.

\bibitem[{Zhu et~al.(2021)Zhu, Wang, Fu, Yan, Narayana, Sone, Basu, and Wang}]{zhu-etal-2021-multimodal}
Zhu, W.; Wang, X.; Fu, T.-J.; Yan, A.; Narayana, P.; Sone, K.; Basu, S.; and Wang, W.~Y. 2021.
\newblock Multimodal Text Style Transfer for Outdoor Vision-and-Language Navigation.
\newblock In \emph{Proceedings of the 16th Conference of the European Chapter of the Association for Computational Linguistics: Main Volume}, 1207--1221. Online: Association for Computational Linguistics.

\end{thebibliography}
\clearpage
\appendix
\section{Finetuning Details}
In the finetuning experiments, we use the official LLaMA-7b weights and finetune a LoRA adapter for \textit{q\_proj} and \textit{v\_proj}. LoRA hyperparameter are set to $r=8$, $alpha=16$, $dropout=0.05$ and no bias. We use Adam~\cite{adam} as the optimizer with a learning rate of $0.0003$, warmup ratio of 0.1 and linear decay. The batch size is 16 and we train for 20 epochs. We use greedy decoding for all experiments.

\section{Modified Environment}
\label{apx:environment}
In Section~\ref{sec:inconsistencies} we propose modifications to the environment introduced by \citet{Chen2018Touchdown}. In Table~\ref{table:comp_intersections} we give an overview of action sequences required to clear 3-way, 4-way and 5-way intersections in different directions in the original environment implementation and our modified environment. It is clear that the action sequences required in our improved environment are more intuitive and are necessary to enable few-short agents to interact with it.

\section{Landmark Extraction}
\label{apx:extraction}
The landmarks mentioned in the navigation instructions are extracted before the run starts. We do this by a separate prompt that we feed to GPT-3. The prompt for Map2seq instructions is shown in Figure~\ref{frame:extraction_map2seq} and the one for Touchdown in Figure~\ref{frame:extraction_touchdown}. It provides five instructions texts paired with a list of extracted landmarks as in-context examples. The lists of example landmarks were compiled by hand and the same prompt is used for each instance. There are no gold annotations for extracted landmarks and as such no quantitative evaluation is possible. In Figure~\ref{frame:extracted_landmarks} we show landmarks extracted by GPT-3 using this prompt.

\section{Landmark Scorer}
\begin{figure}
  \centering
  \begin{subfigure}{0.48\textwidth}
  \includegraphics[width=\textwidth]{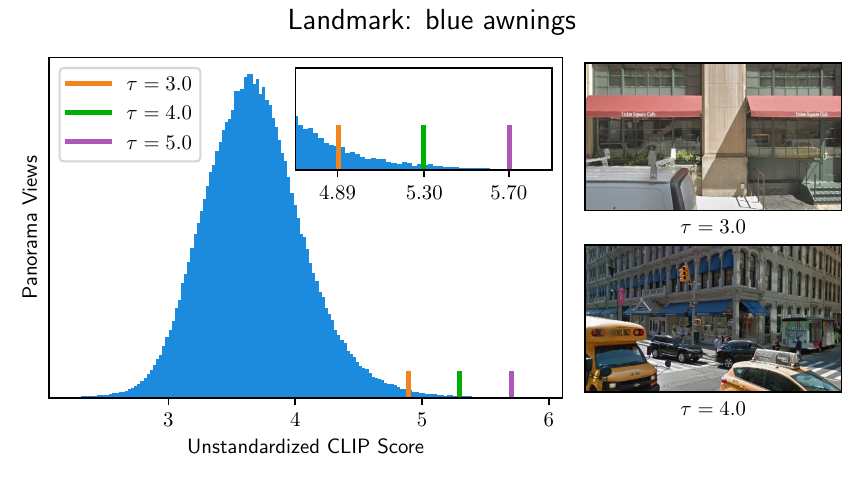}
  \end{subfigure}
  \begin{subfigure}{0.48\textwidth}
  \includegraphics[width=\textwidth]{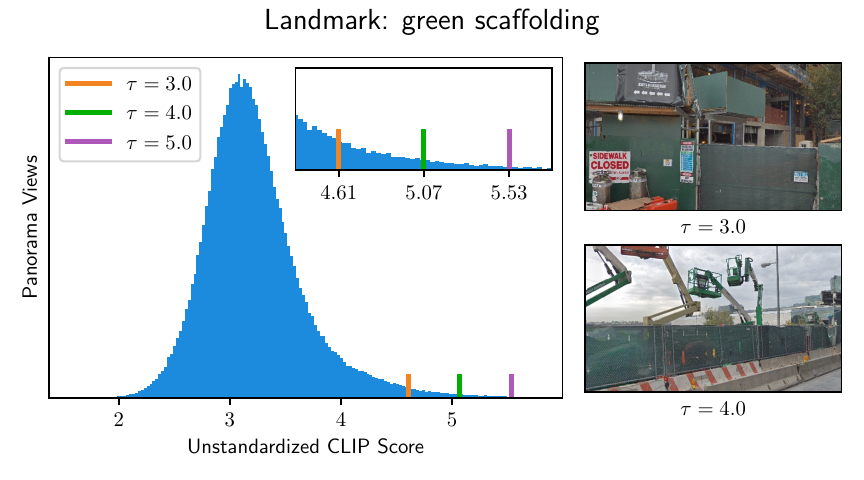}
  \end{subfigure}
  \begin{subfigure}{0.48\textwidth}
  \includegraphics[width=\textwidth]{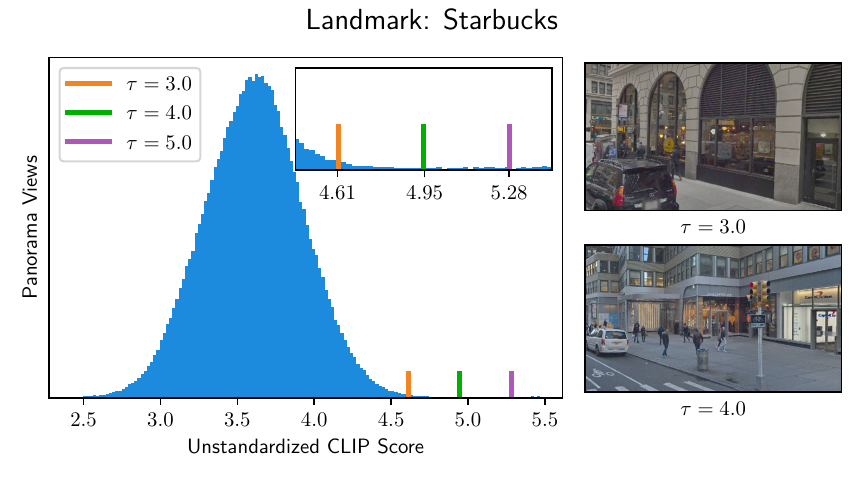}
  \end{subfigure}
  %\begin{subfigure}{0.48\textwidth}
  %\includegraphics[width=\textwidth]{plots/pano_threshold/thresholds_green_wall.pdf}
  %\end{subfigure}
  \caption{Distribution of CLIP scores between a landmark and panorama images in the training area. The CLIP score represents the semantic similarity of the panorama image and the text caption "picture of $[$landmark$]$". The distribution is used to standardize the score of the landmark and a novel panorama. The threshold $\tau$ is defined on the standardized score and used to determine the visibility of the landmark in the novel panorama image.}
  %\vspace{6cm}
  \label{apx:pano_threshold}
\end{figure}

We show the CLIP score distribution and panorama views at certain thresholds for additional landmarks in Figure~\ref{apx:pano_threshold}. Some navigation instructions refer to the flow up traffic when orientating the agent in the beginning of Touchdown instances, e.g. "Orientate yourself with against the flow of traffic...". To support this kind instructions, we score the phrases "the front view of a vehicle" and "the rear view of a vehicle" once, before the start. Whichever phrase scores higher with the initial perspective, determines if the agent is facing against the traffic or with the flow of traffic respectively. This traffic flow prediction is then provided as an environment observation string before the first step of the agent. 

\section{Full Prompt Sequence}
\begin{figure}
    \centering
    \includegraphics[width=0.48\textwidth]{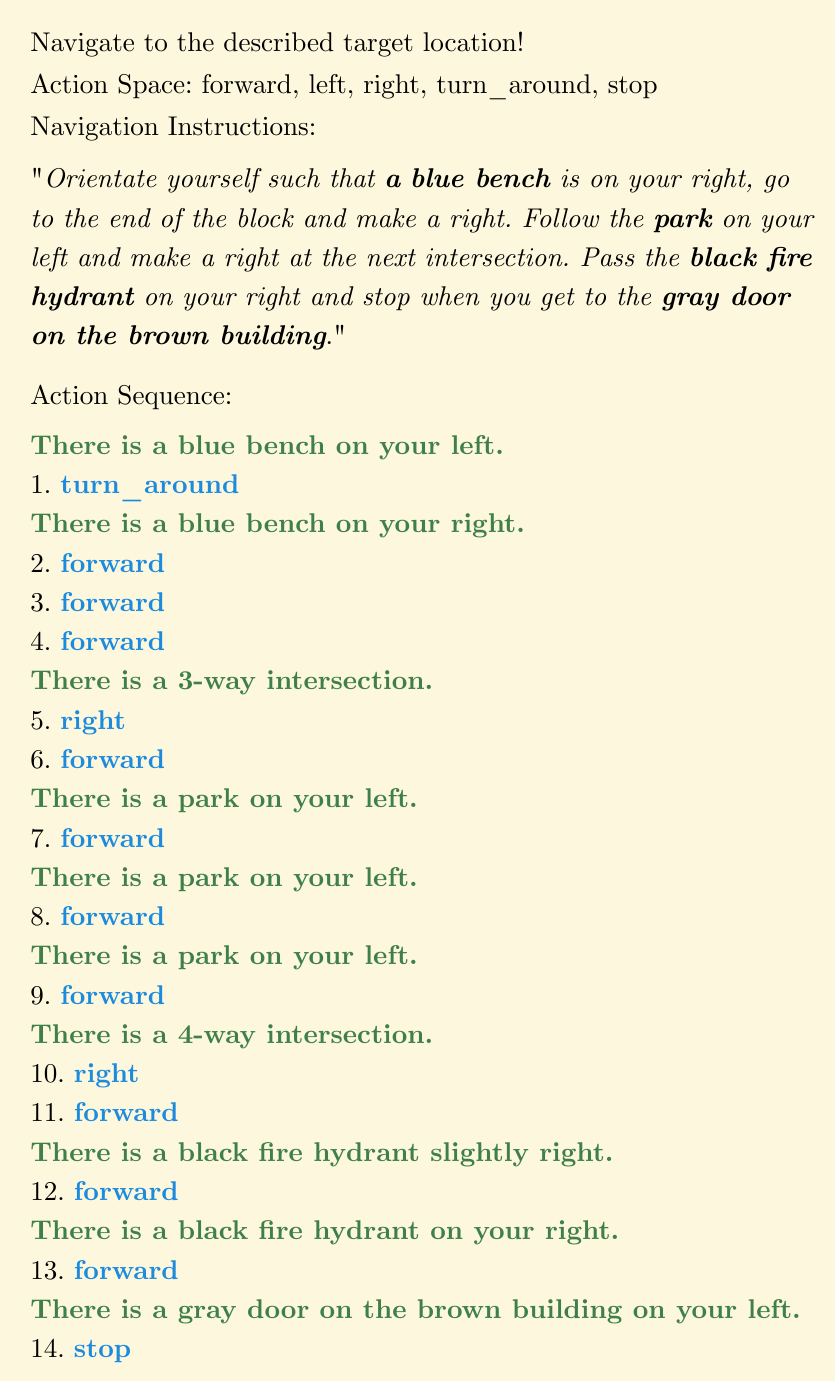}
    \caption{Finished prompt sequence used to utilize LLMs for VLN in Street View. Verbalized observations of the visual environment are in green and appended to the prompt at each step. Agent actions (blue) are acquired by LLM next word prediction. Highlighting of text and shortening of route for visual presentation only. Full navigation trajectories are on average 40 steps long.}
    \label{apx:figure1}
\end{figure}

In Figure~\ref{apx:figure1} we show a full prompt sequence for a given navigation instance. The agent predicted $\act{stop}$ at timestep 14 and thus finished the trajectory. In the depicted case the agent followed the correct route and successfully completed the navigation objective. For visualization purposes the trajectory is shortened. On average the routes in Touchdown and Map2seq require 40 steps to be completed. This also means the agents has to evaluate 200 panorama views for each navigation instance.

\clearpage

\begin{table*}
\centering
%\ra{1.05}
\resizebox{.99\linewidth}{!}{
\begin{tabular}{ccll}
  \toprule
  \textbf{Intersection} & \textbf{Path} & \textbf{Environment by Chen et al.} & \textbf{Our Environment}\\
  \midrule
  \multirow{2}{*}{\fbox{\includegraphics[width=4.3cm]{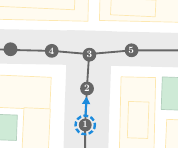}}} & 2\textrightarrow3\textrightarrow4 & $[\act{forward}, \act{left}, \act{forward}]$ & $[\act{forward}, \act{left}, \act{forward}]$\\
  & 2\textrightarrow3\textrightarrow5 & $[\act{forward}, \act{forward}]$ & $[\act{forward}, \act{right}, \act{forward}]$\\
  &  &  & \\
  &  &  & \\
  &  &  & \\
  &  &  & \\
  &  &  & \\
  &  &  & \\
  &  &  & \\
  &  &  & \\
  \midrule
  \multirow{3}{*}{\fbox{\includegraphics[width=4.3cm]{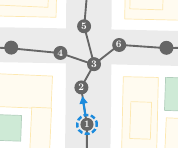}}} & 2\textrightarrow3\textrightarrow4 & $[\act{forward}, \act{left}, \act{left}, \act{forward}]$ & $[\act{forward}, \act{left}, \act{forward}]$\\
  & 2\textrightarrow3\textrightarrow5 & $[\act{forward}, \act{left}, \act{forward}]$ & $[\act{forward}, \act{forward}]$\\
  & 2\textrightarrow3\textrightarrow6 & $[\act{forward}, \act{forward}]$ & $[\act{forward}, \act{right}, \act{forward}]$\\
  &  &  & \\
  &  &  & \\
  &  &  & \\
  &  &  & \\
  &  &  & \\
  &  &  & \\
  &  &  & \\
\midrule
  \multirow{3}{*}{\fbox{\includegraphics[width=4.3cm]{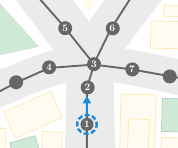}}} & 2\textrightarrow3\textrightarrow4 & $[\act{forward}, \act{left}, \act{left}, \act{forward}]$ & $[\act{forward}, \act{left}, \act{left}, \act{forward}]$\\
  & 2\textrightarrow3\textrightarrow5 & $[\act{forward}, \act{left}, \act{forward}]$ & $[\act{forward}, \act{left}, \act{forward}]$\\
  & 2\textrightarrow3\textrightarrow6 & $[\act{forward}, \act{forward}]$ & $[\act{forward}, \act{right}, \act{forward}]$\\
  & 2\textrightarrow3\textrightarrow7 & $[\act{forward}, \act{right}, \act{forward}]$ & $[\act{forward}, \act{right}, \act{right}, \act{forward}]$\\
  &  &  & \\
  &  &  & \\
  &  &  & \\
  &  &  & \\
  &  &  & \\
  &  &  & \\
  \bottomrule
\end{tabular}
}
\caption{Comparison of the Touchdown environment implemented by \citet{Chen2018Touchdown} and the improved implementation proposed by us. The action sequence required to clear an intersection in different directions in our improved environment is semantically aligned with the expected outcome.}
\label{table:comp_intersections}
\end{table*}

\begin{figure*}[ht]
\begin{mdframed}%[frametitle={Prompt:}]
\small
Head to the end of the block and make a right. Pass a Subway entrance on the right and go through the light. At the next light with Staples on the corner, make a right. Stop in front of the library that is a few buildings down on the right.\\
Landmarks:\\
1. a subway entrance\\
2. Staples\\
3. a library\\\\
Go straight through the light ahead of you, then turn right at the next one. After your turn, you will see Starbucks on the left. At the light after that, turn left. Pass the church on the left and then stop after Hot Kitchen. You should be able to see a bike rental on the right.\\
Landmarks:\\
1. Starbucks\\
2. a church\\
3. Hot Kitchen\\
4. a bike rental\\\\
Head to the intersection and turn left. Continue to the end of the block and turn right. Go straight and past the intersection. Stop 1/3 of the way down the block with the large building on your right.\\
Landmarks:\\
None\\\\
Walk to the light with Just Sweet and turn right. Go through a light with an AMC and a couple more blocks until you see a tiny park or plaza on the far left corner. Turn left passing that park and then make a left turn almost immediately after. Stop after a couple of steps, where a road from the right joins the main road.\\
Landmarks:\\
1. Just Sweet\\
2. AMC\\
3. a park\\
4. a plaza\\\\
Go straight through the next 3 lights past the bus stops and at the 4th light shortly after the 3rd take a left. Stop just past the bus stop and Neta diner.\\
Landmarks:\\
1. bus stop\\
2. Neta diner\\\\
\{\textit{navigation instructions}\}\\
Landmarks:\\
\textless \textgreater
\end{mdframed}
\caption{Prompt to extract landmarks from navigation instruction in Map2seq.}
\label{frame:extraction_map2seq}
\end{figure*}
\begin{figure*}[ht]
\begin{mdframed}%[frametitle={Prompt:}]
\small
You will start of at an intersection. To begin, make sure you are going in the direction of the blue and white van with orange cones around it. Pass that van. Go straight through the first intersection you get to. You will come to a light at an intersection where there is a building with a green awning. Take a right. Go straight until you are in the middle of the intersection. In front of you, there is a building with a red sign above the entrance.\\
Landmarks:\\
1. a blue and white van\\
2. orange cones\\
3. a green awning\\
4. a red sign above the entrance\\\\
Turn to the right until you're looking down the street. There should be a red SUV on the right side of the frame now. Begin moving forward until you reach an intersection. Take a left here. Keep moving forward until reaching a three-way intersection. Take another left here. Move forward three times. Turn to the right until you see a red and white street sign next to a series of green boards.\\
Landmarks:\\
1. a red SUV\\
2. a red and white street sign\\
3. a series of green boards\\\\
Head in the direction of traffic and continue going straight. You will have the opportunity to turn right, but DON'T. Keep going straight. When you reach the intersection, turn left. Keep going straight. You will reach an intersection, but keep going straight. Just before you reach the next intersection, you will see a bus stop on the right in front of a credit union.\\
Landmarks:\\
1. a bus stop\\
2. credit union\\\\
If you look around there should be a beige building on your right and a green awning. You want to head in the same direction as the the red building with a staircase and a green awning if you check your surrounding. Make a left turn at the intersection when you arrive. Follow the road until you reach another intersection. At this intersection make a left turn. You should be in an alley. If you go up a few steps there should be a bicycle leaning on a tree. There should be a white car next to the bike. Up ahead at least one step is a silver car and a light green car.\\
Landmarks:\\
1. beige building\\
2. green awning\\
3. a red building with a staircase and a green awning\\
4. a bicycle leaning on a tree\\
5. a white car next to the bike\\
6. a silver car\\
7. a light green car\\\\
Turn so your facing the intersection. You will take one step and be in the intersection. Turn Left, you will see some construction barriers on your left. Go one block and at the very next intersection go left again. Go about half a block or so and you will see another orange barricade on your left. There will be some tarps covering construction stuff and scaffolding. At the beginning of the barricade, there is an orange safety light.\\
Landmarks:\\
1. construction barriers\\
2. orange barricade\\
3. tarps covering construction stuff and scaffolding\\
4. orange safety light\\\\
\{\textit{navigation instructions}\}\\
Landmarks:\\
\textless \textgreater
\end{mdframed}
\caption{Prompt to extract landmarks from navigation instruction in Touchdown.}
\label{frame:extraction_touchdown}
\end{figure*}
\begin{figure*}[ht]
\begin{mdframed}

\textbf{Map2seq:}
\small
\begin{mdframed}[frametitle={Navigation Instructions (ID: 6197):}]
Head through the first intersection and at the next light make a right. Go past the next light and the Butcher Daughter will be on the far left corner. At the next light make a left and stop in front of Kings Avenue Tattooing.\\\\
\textbf{Extracted Landmarks:} "The Butcher Daughter", "Kings Avenue Tattooing"
\end{mdframed}

\begin{mdframed}[frametitle={Navigation Instructions (ID: 6205):}]
Head past the market and the cathedral and make a right at the light. At the next light with the Delicatessen on the corner make a left. Stop in front of the fire hall.
\\\\
\textbf{Extracted Landmarks:} "a market", "a cathedral", "a Delicatessen", "a fire hall"
\end{mdframed}

\begin{mdframed}[frametitle={Navigation Instructions (ID: 6211):}]
Go to the end of the block and turn left. Pass More Parlour on the right and turn right at the lights. Go past the park on the left to the lights and turn left and take two steps. Stop at Straus Square on the right before the bike rental.
\\\\
\textbf{Extracted Landmarks:} "More Parlour, "a park", "Straus Square", "a bike rental"
\end{mdframed}

\begin{mdframed}[frametitle={Navigation Instructions (ID: 6227):}]
Turn right at the lights. Pass Spitzer's Corner on the next left and turn left. Go down the long block and through the double set of lights. Stop just before Farmhouse on the right corner.
\\\\
\textbf{Extracted Landmarks:} "Spitzer's Corner", "Farmhouse"
\end{mdframed}

\vspace{0.3cm}
\normalsize
\textbf{Touchdown:}
\small
\begin{mdframed}[frametitle={Navigation Instructions (ID: 546):}]
You're going to go down the narrow street, not the big/main street here. Turn yourself so you've got that big mural of a guy with nunchucks at your back, and you're facing down the narrow street where you'll go in the same direction the parked cars are facing. Go down that street, and pass through the first intersection with the stop sign. At the second intersection, turn right. Go until you're nearly in the next intersection (right before you'd be standing on the crosswalk).
\\\\
\textbf{Extracted Landmarks:} "mural of a guy with nunchucks", "parked cars", "stop sign", "crosswalk"
\end{mdframed}

%\begin{mdframed}[frametitle={Navigation Instructions (ID: 553):}]
%Stand so that you are walking with the flow of traffic. Walk to the intersection and turn right. Walk straight, passing through the first intersection. You will see a purple awning on your right. Just past it is a black awning.
%\\\\
%\textbf{Extracted Landmarks:} "purple awning", "black awning"
%\end{mdframed}

\begin{mdframed}[frametitle={Navigation Instructions (ID: 580):}]
You're basically starting in an intersection. Move to the center of the intersection, and turn yourself so the restaurant with the bright yellow awnings and sidewalk barriers is on your right side (you'll pass it on your right as you walk down the street). Go down that street, with the yellow restaurant on your right, and go to the next intersection. Turn right. Look at the buildings on your right. A short way down the block you'll come to a bar with a wood bench out front. There is also a red velvet rope near the bench.
\\\\
\textbf{Extracted Landmarks:} "restaurant with bright yellow awnings and sidewalk barriers", "bar with a wood bench", "red velvet rope"
\end{mdframed}

\begin{mdframed}[frametitle={Navigation Instructions (ID: 584):}]
Turn yourself around left so that you are going with the flow of traffic, there should be a green door on your right. Go forward and make a right turn at the first intersection. There will be a black awning on your right. Continue forward. When you come to the next intersection, make another right turn. As you get near the next intersection, you will see large red brick buildings on your right. You will see a pallet of green sandbags sitting along the sidewalk.
\\\\
\textbf{Extracted Landmarks:} "green door black awning", "large red brick buildings", "pallet of green sandbags"
\end{mdframed}

\end{mdframed}
\caption{Landmarks extracted by GPT-3 using the 5-shot prompt for Map2seq and Touchdown.}
\label{frame:extracted_landmarks}
\end{figure*}

\end{document}